\documentclass{article}

\usepackage{microtype}
\usepackage{graphicx}
\usepackage{subcaption}
\usepackage{booktabs} 

\usepackage{hyperref}

\usepackage[preprint]{icml2026}

\usepackage{amsmath}
\usepackage{amssymb}
\usepackage{mathtools}
\usepackage{amsthm}
\newcommand{\cmark}{\ensuremath{\checkmark}}
\newcommand{\xmark}{\ensuremath{\times}}
\usepackage[table]{xcolor}
\usepackage{placeins}
\usepackage{multirow}
\usepackage{booktabs}
\usepackage{multirow}
\usepackage{pifont} 

\usepackage[capitalize,noabbrev]{cleveref}

\theoremstyle{plain}

\theoremstyle{definition}

\theoremstyle{remark}

\usepackage[textsize=tiny]{todonotes}

\icmltitlerunning{Efficient Post-Training Pruning
of Large Language Models with Statistical Correction}

\begin{document}

\twocolumn[
 \icmltitle{Efficient Post-Training Pruning\\
of Large Language Models with Statistical Correction}





  \icmlsetsymbol{equal}{*}

\begin{icmlauthorlist}
  \icmlauthor{Peiqi Yu}{scu}
  \icmlauthor{Jinhao Wang}{scu}
  \icmlauthor{Xinyi Sui}{scu}
  \icmlauthor{Nam Ling}{scu}
  \icmlauthor{Wei Wang}{futurewei}
  \icmlauthor{Wei Jiang}{futurewei}
\end{icmlauthorlist}

\icmlaffiliation{scu}{Department of Computer Science and Engineering, Santa Clara University, Santa Clara, CA, USA}
\icmlaffiliation{futurewei}{Futurewei Technologies, Inc., USA}

\icmlcorrespondingauthor{Peiqi Yu}{pyu@scu.edu}

\printAffiliationsAndNotice{\icmlEqualContribution}

  \icmlkeywords{Machine Learning, ICML}

  \vskip 0.3in
]



\printAffiliationsAndNotice{}  

\begin{abstract}
Post-training pruning is an effective approach for reducing the size and inference cost of large language models (LLMs), but existing methods often face a trade-off between pruning quality and computational efficiency. Heuristic pruning methods are efficient but sensitive to activation outliers, while reconstruction-based approaches improve fidelity at the cost of heavy computation. In this work, we propose a lightweight post-training pruning framework based on first-order statistical properties of model weights and activations. During pruning, channel-wise statistics are used to calibrate magnitude-based importance scores, reducing bias from activation-dominated channels. After pruning, we apply an analytic energy compensation to correct distributional distortions caused by weight removal. Both steps operate without retraining, gradients, or second-order information. Experiments across multiple LLM families, sparsity patterns, and evaluation tasks show that the proposed approach improves pruning performance while maintaining computational cost comparable to heuristic methods. The results suggest that simple statistical corrections can be effective for post-training pruning of LLMs.


\end{abstract}

\section{Introduction}

Large Language Models (LLMs) \cite{openai2024gpt4technicalreport,brown2020languagemodelsfewshotlearners,llama,llama2} have achieved strong performance across a broad range of language understanding and generation tasks \cite{bommarito2022gpttakesbarexam,wei2022emergentabilitieslargelanguage,bubeck2023sparksartificialgeneralintelligence,devlin2019bertpretrainingdeepbidirectional,hendrycks2021measuringmassivemultitasklanguage}. However, multiple lines of work, such as pretraining language modeling \cite{frankle2019lotterytickethypothesisfinding}, machine translation \cite{voita2023neuronslargelanguagemodels}, and downstream fine-tuning \cite{sanh2020movementpruningadaptivesparsity}, have consistently shown that these models are substantially over-parameterized \cite{gale2019statesparsitydeepneural,zhu2017prunepruneexploringefficacy}. In particular, large fractions of model parameters can be pruned or compressed with minimal impact on performance, even without additional retraining \cite{frantar2023gptqaccurateposttrainingquantization,sparsegpt,lin2024awqactivationawareweightquantization}. Exploiting this redundancy through post-training model reduction is therefore important for reducing computational, memory, and storage costs, and for enabling scalable deployment of LLMs across diverse hardware and application settings \cite{sparsegpt,mishra2021acceleratingsparsedeepneural,hoffmann2022trainingcomputeoptimallargelanguage,strubell2019energypolicyconsiderationsdeep,kaplan2020scalinglawsneurallanguage}.

Post-training model reduction for LLMs involves a trade-off between pruning accuracy and computational cost. Post-training methods can be divided into reconstruction-based approaches that model weight interactions and heuristic approaches based on simple importance metrics \cite{LeCun1990OBD,HassibiOBS1993}. Reconstruction-based methods such as SparseGPT \cite{sparsegpt} leverage second-order information to update the remaining weights after pruning to preserve reconstruction quality, at the cost of increased computational and memory overhead from Hessian estimation and matrix inversion. Heuristic approaches such as Wanda \cite{wanda} avoid weight updates and instead perform greedy pruning based on weight–activation magnitudes, which improves scalability. However, these methods analyze weights in isolation and rely  on input signal strength to assess parameter importance, resulting in suboptimal pruning decisions \cite{magnitude,ria,dettmers2022,frankle2021pruningneuralnetworksinitialization}. In particular, channels dominated by extreme activation outliers may receive high importance scores despite their instability, as shown in Sec.~\ref{variancecal}.

\begin{figure}[t]
  \centering
  \includegraphics[width=\columnwidth]{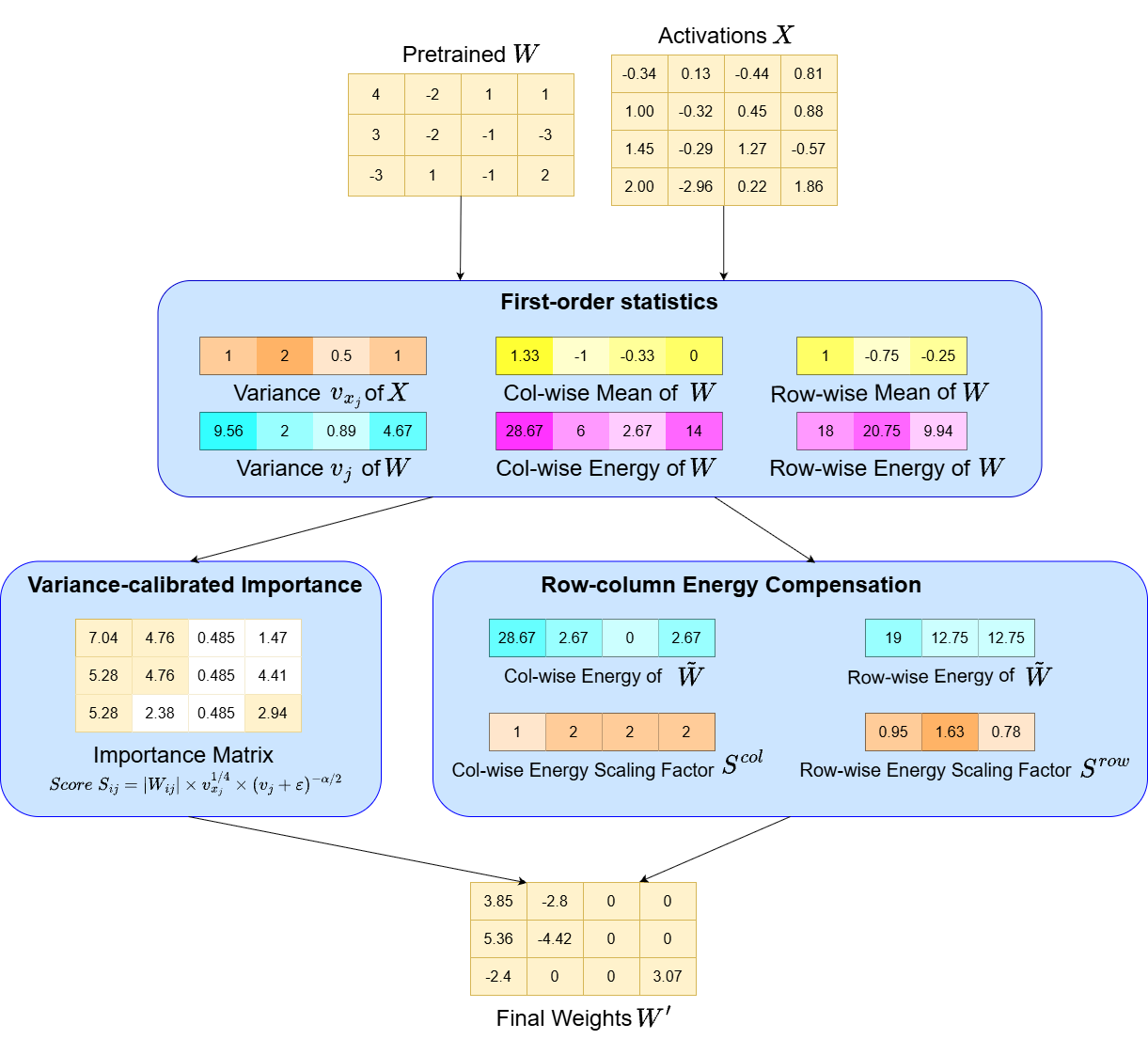}
  \caption{
Overview of the proposed statistical correction framework.
First-order statistics from pretrained weights $W$ and activations $X$ are used for variance-calibrated importance scoring and post-pruning energy compensation.
The two components are applied independently and produce corrected weights while preserving the sparsity pattern.
  }
  \label{fig:method_overview}
\end{figure}

Recent work has also explored alternative approaches for defining parameter importance. For example, DenoiseRotator \cite{denoiserotator} applies orthogonal transformations to redistribute importance scores before pruning, while RIA \cite{ria} adopts a weight-pruning criterion based on relative importance among parameters. These methods rely on preprocessing or weight-only importance estimation and do not explicitly model activation reliability or post-pruning signal correction. In contrast, we focus on post-training pruning and weight adjustment directly on pretrained models.

We propose a post-training pruning framework based on first-order statistical properties of both model weights and activations. During pruning, channel-wise statistics of weights and activations are used to calibrate magnitude-based importance scores, reducing bias from activation-dominated channels \cite{wanda}. After pruning, signal distortion caused by weight removal is corrected through an analytic energy compensation that restores the scale of layer outputs. Both parameter selection and weight adjustment rely on first-order statistical correction and do not require gradient-based optimization or second-order information. As a result, the method has computational cost comparable to heuristic pruning approaches \cite{magnitude,wanda} while avoiding reconstruction-based overhead. Moreover, the proposed weight adjustment is criterion-agnostic and can be applied after pruning regardless of the underlying importance metric, making it compatible with existing weight-pruning methods.

For pruning, we introduce a variance-calibrated importance metric that corrects systematic bias in magnitude-based heuristics. Existing methods such as Wanda \cite{wanda} compute importance scores using the product of weight magnitude and input activation norm ($|W_{ij}| \cdot \|X_j\|_2$). This formulation can be overly sensitive to activation outliers \cite{dettmers2022,kovaleva2021bertbustersoutlierdimensions}. Channels with large input norms tend to amplify the importance scores of all associated weights, including noisy or structurally insignificant parameters.

To address this issue, we incorporate weight variance as a measure of channel stability. Feature channels with high weight variance tend to exhibit heterogeneous and unstable distributions \cite{ria,voita2023neuronslargelanguagemodels,kovaleva2021bertbustersoutlierdimensions,timkey2021barkbiteroguedimensions}, making magnitude-based importance scores less reliable. In contrast, low-variance channels reflect more consistent parameter structure. We therefore down-weight activation-driven importance scores using an inverse variance factor, which effectively suppresses activation-dominated channels and reallocates sparsity toward distributionally stable features.

For weight adjustment, we observe that a major source of post-pruning degradation is mismatched signal energy rather than irrecoverable information loss. Pruning alters the aggregate signal strength of feature maps, leading to a distribution shift that accumulates across layers \cite{magnitude,li2016revisiting,sun2016deep}. To address this effect, we introduce an energy-based compensation mechanism that restores the scale of latent representations after pruning.

Specifically, we apply a deterministic, closed-form rescaling that aligns the layer-wise output energy of the pruned model with that of the original model \cite{HassibiOBS1993,LeCun1990OBD}. This update is non-iterative and requires no gradient computation or second-order information \cite{sparsegpt,wanda}. By restoring the dynamic range of intermediate activations, the proposed compensation alleviates pruning-induced distortion while retaining computational complexity comparable to heuristic pruning methods.

Our contributions are summarized as follows:
\begin{itemize}
    \item We propose a post-training pruning framework that uses first-order statistical correction for both parameter selection and post-pruning weight adjustment.
    \item We introduce a variance-calibrated importance metric that corrects activation-induced bias in magnitude-based pruning without relying on gradient or second-order information.
    \item We develop an effective energy compensation method that restores activation scale after pruning and can be applied independently of the pruning criterion, with computational cost comparable to heuristic approaches in practice.
    \item We evaluate the proposed method on multiple LLMs and sparsity settings, achieving consistent improvements over existing post-training pruning methods \cite{llama,llama2,llama3,qwen25}.
\end{itemize}

\section{Preliminaries}\label{sec2}

Post-training pruning is a widely used approach for compressing LLMs without the high cost of retraining \cite{sparsegpt,wanda}.
We consider a pretrained LLM composed of stacked linear layers. Let $W \in \mathbb{R}^{d_{\text{out}}\times d_{\text{in}}}$
denote the pretrained weight matrix of a given linear layer. 
Post-training pruning methods construct a binary mask $M \in \{0,1\}^{d_{\text{out}}\times d_{\text{in}}}$, where $M_{ij}=0$ indicates that the corresponding weight is pruned.
By applying the mask the pruned weight matrix can be obtained as $\widetilde{W} = M\odot W$. 

\paragraph{\textbf{Heuristic-based weight pruning}}

Early magnitude-based methods \cite{magnitude,LeCun1990OBD} prune weights with small absolute values under the assumption that weight importance is independent of input data. More recent approaches incorporate activation statistics to better capture feature importance. For example, Wanda \cite{wanda} computes importance scores using the product of weight magnitude and input activation norm, $|W_{ij}| \cdot \|X_j\|_2$. However, activation magnitudes in LLMs can be highly heterogeneous \cite{dettmers2022,xiao2023,kovaleva2021bertbustersoutlierdimensions}, making such heuristics sensitive to activation outliers. As a result, structurally insignificant weights may receive high importance scores due to amplification by a small number of dominant input features \cite{timkey2021barkbiteroguedimensions}.

\paragraph{\textbf{Reconstruction-based weight adjustment}}

To address performance degradation after pruning, reconstruction-based methods further adjust the remaining weights in $\widetilde{W}$. Methods such as SparseGPT \cite{sparsegpt} and its variants \cite{ma2023llmpruner, vanderouderaa2024llmsurgeon,frantar2023gptqaccurateposttrainingquantization,frantar2023optimalbraincompressionframework,hubara2021acceleratedsparseneuraltraining} formulate pruning as a reconstruction problem and update the retained weights to minimize the error between the original and pruned layer outputs. These methods typically rely on second-order information, using approximations of the Hessian to guide weight updates. While effective, the required matrix factorizations and iterative updates cause significant computational and memory overhead. 

\section{Method}

We present a post-training pruning method based on first-order activation statistics that addresses two limitations of previous approaches \cite{sparsegpt,wanda,magnitude}: activation-induced bias in magnitude-based weight selection and signal energy mismatch after pruning. The method handles both issues directly using closed-form statistical quantities, without gradient-based optimization or second-order information.

\subsection{Variance-Calibrated Weight Selection}\label{variancecal}

Magnitude-based heuristic pruning methods such as Wanda \cite{wanda} are sensitive to activation outliers \cite{dettmers2022,kovaleva2021bertbustersoutlierdimensions}, which can inflate importance scores for unstable weight columns. We address this effect using Column-Variance Reweighting (CVR), which calibrates column-wise importance scores by weight variance and rectified channel activation.


Specifically, given the weight matrix $W \in \mathbb{R}^{d_{\text{out}} \times d_{\text{in}}}$ and its input activation $X$, feature channels with high weight variance tend to exhibit heterogeneous and unstable distributions, while activation outliers can be directly characterized through activation dispersion. 

Therefore, based on weight $W$, we compute the variance of each input column as:
\[
v_j = \mathrm{Var}_{i=1}^{d_{\text{out}}}(W_{ij}),
\]
and define a weight-based calibration factor:
\[
c_j = (v_j + \varepsilon)^{-\alpha/2},
\]
where $\alpha \ge 0$ controls the reweighting strength and $\varepsilon$ is a small constant for numerical stability.

At the same time, based on activation $X$, we estimate the channel-wise activation variance:
\[
v_{x_j}  = \mathbb{E}[x_j^2] - \mathbb{E}[x_j]^2,
\]
and compute an activation-based calibration factor:
\[
a_j^{(\mathrm{var})} = v_{x_j}^{1/4}.
\]

The proposed CVR rescales the column-wise importance scores using these calibration factors and the final weight importance is given by:
\[
S_{ij} = |W_{ij}| \cdot a_j \cdot c_j.
\]

In contrast, Wanda defines a pruning metric based on per-channel activation statistics derived from the second moment, corresponding to an activation factor of the form $a_j\!\propto\!\sqrt{\mathbb{E}[x_j^2]}$. This can amplify all weights in channels with large activation variance, including insignificant parameters.  
By explicitly accounting for activation variance and incorporating weight variance through $c_j$, CVR suppresses activation-dominated channels and reduces false amplification. Since $c_j$ is computed solely from pretrained weights, CVR has no additional activation or gradient computation.

\subsection{Energy-Compensated Weight Adjustment}

After pruning, the mask $M$ is fixed and the remaining weights in $\tilde{W}=M\odot W$ are adjusted to obtain corrected weights $W'$. Pruning changes the aggregate signal strength of weight vectors, causing distributional distortions that can accumulate across layers. Two effects are particularly relevant \cite{sparsegpt,xiao2023}.

\paragraph{Mean shift.}
Removing a subset of weights disrupts the balance between positive and negative contributions, making offset of weight vectors sensitive to sparsity patterns.

\paragraph{Energy collapse.}
Pruning reduces the centered $\ell_2$ energy of weight vectors, attenuating the magnitude of signals propagated through the network even when structurally important weights are retained \cite{dettmers2022}.

Existing reconstruction-based methods such as SparseGPT adjust pruned weights through reconstruction objectives, without explicitly correcting these distributional distortions. In contrast, we apply a lightweight post-pruning correction that matches the centered energy of the pruned weights to that of the original pretrained weights, using the original mean as a stable reference. The correction is applied independently along the input (column-wise) and output (row-wise) dimensions.

\paragraph{Intuition.}
Column-wise correction compensates for energy loss in pruned input connections, restoring the scale of signals entering each output unit. Row-wise correction compensates for energy loss in outgoing connections, restoring the scale of signals produced by each output unit. Applying both preserves the overall signal balance of the layer.

\paragraph{Column-wise correction.}
Let
\[
\mu^{\text{col}}_j=\frac{1}{d_{\text{out}}}\sum\nolimits_{i=1}^{d_{\text{out}}}W_{ij}
\]
denote the column-wise mean of the original weights. Using this mean for centering, the centered column-wise energies are defined as:
\[
E^{\text{col}}_{\text{orig}}(j)=\sum_{i=1}^{d_{\text{out}}}(W_{ij}-\mu^{\text{col}}_j)^2,
E^{\text{col}}_{\text{pruned}}(j)=\sum_{i=1}^{d_{\text{out}}}(\tilde{W}_{ij}-\mu^{\text{col}}_j)^2.
\]
We apply the transformation:
\[
\tilde{W}\leftarrow(\tilde{W}-\mu^{\text{col}})\odot s^{\text{col}}+\mu^{\text{col}},
s^{\text{col}}_j=\sqrt{\frac{E^{\text{col}}_{\text{orig}}(j)}{E^{\text{col}}_{\text{pruned}}(j)+\varepsilon}},
\]
where $\varepsilon$ is a small constant for numerical stability. Scaling factors $s^{\text{col}}_j$ are clamped to a fixed range to avoid over-amplification in highly sparse columns, and the pruning mask is re-applied to preserve exact sparsity.

\paragraph{Row-wise correction.}
A similar procedure is applied along the output dimension. Let
\[
\mu^{\text{row}}_i=\frac{1}{d_{\text{in}}}\sum\nolimits_{j=1}^{d_{\text{in}}}W_{ij}
\]
denote the row-wise mean of the original weights, with centered energies defined as:
\[
E^{\text{row}}_{\text{orig}}(i)=\sum_{j=1}^{d_{\text{in}}}(W_{ij}-\mu^{\text{row}}_i)^2,
E^{\text{row}}_{\text{pruned}}(i)=\sum_{j=1}^{d_{\text{in}}}(\tilde{W}_{ij}-\mu^{\text{row}}_i)^2.
\]
The corrected weights are obtained as:
\[
W'=(\tilde{W}-\mu^{\text{row}})\odot s^{\text{row}}+\mu^{\text{row}},
s^{\text{row}}_i=\sqrt{\frac{E^{\text{row}}_{\text{orig}}(i)}{E^{\text{row}}_{\text{pruned}}(i)+\varepsilon}},
\]
and the pruning mask is re-applied. In practice, column-wise correction is applied first, followed by row-wise correction.

Because the proposed compensation depends only on post-pruning weights and closed-form statistics, it is independent of the pruning criterion and can be applied after mask construction, making it compatible with existing weight-pruning methods.

\subsection{Algorithm Summary}

Algorithm~\ref{alg:ptp} summarizes the proposed post-training pruning pipeline.
The method proceeds in two steps.
In the first stage, it collects first-order activation statistics on a pretrained model and applies variance-calibrated importance scoring (Sec.~3.1) to construct the pruning mask \cite{sparsegpt,wanda}.
In the second stage, it performs mean-shifted energy matching (Sec.~3.2) to correct pruning-induced distortion in the remaining weights using closed-form updates \cite{HassibiOBS1993}.


\renewcommand{\algorithmicrequire}{\textbf{Input:}}
\renewcommand{\algorithmicensure}{\textbf{Output:}}

\begin{algorithm}[htb]
\caption{Post-Training Pruning with Variance-Calibrated Selection and Energy Matching}
\label{alg:ptp}
\begin{algorithmic}[1]
\REQUIRE Pretrained weights $W$, calibration data $\mathcal{D}$ (for mask construction), target sparsity (or $n\!:\!m$ pattern)
\ENSURE Sparse corrected weights $W'$\vspace{.3em}

\STATE \textbf{Procedure: Mask construction (Sec.~3.1)}
\STATE Collect activation statistics on $\mathcal{D}$ and compute activation calibration factors $\{a_j\}$.
\STATE Compute weight variance calibration factors $\{c_j\}$.
\STATE Compute importance scores $S_{ij} = |W_{ij}|\cdot a_j \cdot c_j$ and construct pruning mask $M$.
\STATE \textbf{Procedure: Post-pruning adjustment (Sec.~3.2)}
\STATE Apply mask: $\widetilde{W} \leftarrow M \odot W$.
\STATE Apply column-wise correction to $\widetilde{W}$.
\STATE Apply row-wise correction to $\widetilde{W}$ and obtain $W'$.
\end{algorithmic}
\end{algorithm}

\begin{table*}[!t]
\caption{Perplexity on WikiText-2 under different sparsity settings. Lower is better.}
\label{tab:wikitext_ppl_main}
\centering
\begin{small}
\setlength{\tabcolsep}{4pt}
\renewcommand{\arraystretch}{1.15}
\begin{tabular}{l c c ccc cc cccc}
\toprule
\multirow{2}{*}{Method} & \multirow{2}{*}{Weight Update} & \multirow{2}{*}{Sparsity}
& \multicolumn{3}{c}{LLaMA-2}
& \multicolumn{2}{c}{LLaMA-3}
& \multicolumn{4}{c}{Qwen2.5} \\
\cmidrule(lr){4-6} \cmidrule(lr){7-8} \cmidrule(lr){9-12}
& & & 7B & 13B & 70B & 8B & 70B & 7B & 14B & 32B & 72B \\
\midrule
Pretrained Dense & -- & 0
& 5.12 & 4.57 & 3.12 & 6.14 & 2.86 & 6.85 & 5.29 & 5.02 & 3.88 \\
\midrule
\rowcolor{gray!10}
Magnitude & No & 50\%
& 14.90 & \textbf{6.37} & \textbf{4.98} & \textbf{310.68} & 10.58 & 395.41 & 22.94 & \textbf{19.22} & 734.04 \\
\rowcolor{gray!10}
Magnitude+EC & Yes & 50\%
& \textbf{11.56} & 7.04 & 5.73 & 735.39 & \textbf{9.03} & \textbf{174.24} & \textbf{20.14} & 21.49 & \textbf{112.79} \\
\rowcolor{blue!10}
Wanda & No & 50\%
& 6.41 & 5.46 & 3.98 & 9.86 & \textbf{5.80} & 10.60 & 8.24 & 6.30 & 5.22 \\
\rowcolor{blue!10}
Wanda+EC & Yes & 50\%
& \textbf{6.27} & \textbf{5.46} & \textbf{3.86} & \textbf{8.22} & 6.83 & \textbf{7.75} & \textbf{7.30} & \textbf{6.00} & \textbf{4.80} \\
\rowcolor{gray!20}
Ours (CVR+EC) & Yes & 50\%
& \textbf{6.19} & \textbf{5.32} & \textbf{3.81} & \textbf{8.17} & \textbf{5.18} & \textbf{7.67} & \textbf{6.78 }& \textbf{5.91} & \textbf{4.79} \\
\midrule
\rowcolor{gray!10}
Magnitude & No & 4:8
& 16.53 & 6.76 & \textbf{5.54} & \textbf{207.82} & 10.73 & 2366.09 & 27.70 & 21.36 & 456.80 \\
\rowcolor{gray!10}
Magnitude+EC & Yes & 4:8
& \textbf{12.17} & \textbf{6.63} & 5.75 & 283.26 & \textbf{9.26} & \textbf{424.20} & \textbf{22.14} & \textbf{19.09} & \textbf{178.07} \\
\rowcolor{blue!10}
Wanda & No & 4:8
& 7.51 & 6.14 & 4.47 & 12.24 & \textbf{7.89} & 19.49 & 19.79 & 7.07 & 5.93 \\
\rowcolor{blue!10}
Wanda+EC & Yes & 4:8
& \textbf{7.13} & \textbf{6.03} & \textbf{4.27} & \textbf{10.60} & 13.26 & \textbf{8.67} & \textbf{7.85} & \textbf{6.52} & \textbf{5.35} \\
\rowcolor{gray!20}
Ours (CVR+EC) & Yes & 4:8
& \textbf{7.05} & \textbf{5.86} & \textbf{4.24} & \textbf{10.47} & \textbf{7.74} & \textbf{8.56} & \textbf{7.60} & \textbf{6.42} & \textbf{5.33} \\
\midrule
\rowcolor{gray!10}
Magnitude & No & 2:4
& 141.96 & 18.17 & \textbf{6.33} & 3202.50 & \textbf{18.17} & 4119.33 & 58.93 & 24.27 & 287.70 \\
\rowcolor{gray!10}
Magnitude+EC & Yes & 2:4
& \textbf{20.68} & \textbf{7.63} & 6.56 & \textbf{1111.14} & 34.54 & \textbf{522.03} & \textbf{31.24} & \textbf{18.59} & \textbf{158.27} \\
\rowcolor{blue!10}
Wanda & No & 2:4
& 9.51 & 7.31 & 5.16 & 25.19 & \textbf{9.39} & 61.50 & 53.41 & 8.08 & 6.69 \\
\rowcolor{blue!10}
Wanda+EC & Yes & 2:4
& \textbf{8.35} & \textbf{6.80} & \textbf{4.68} & \textbf{15.53} & 54.02 & \textbf{9.90} & \textbf{8.65} & \textbf{7.11} & \textbf{5.89} \\
\rowcolor{gray!20}
Ours (CVR+EC) & Yes & 2:4
& \textbf{8.17} & \textbf{6.50} & \textbf{4.63} & 16.40 & 12.00 & \textbf{9.67} & \textbf{8.43} & \textbf{7.02} & \textbf{5.86} \\
\bottomrule
\end{tabular}
\end{small}
\end{table*}

\section{Experiment}

\subsection{Experimental Setup}

\paragraph{Models.}
We evaluate our method on three representative LLM families: LLaMA-2 \cite{llama2}, LLaMA-3 \cite{llama3}, and Qwen2.5 \cite{qwen25}, covering a range of architectures and model scales. Specifically, we report results on LLaMA-2 (7B, 13B, 70B), LLaMA-3 models, and Qwen2.5 models selected to closely match these parameter scales.

\paragraph{Pruning Methods.}
We compare the proposed method against widely used post-training pruning baselines, including magnitude pruning~\cite{magnitude}, Wanda~\cite{wanda}, and SparseGPT~\cite{sparsegpt}. All baseline methods are evaluated using their original implementations and recommended settings.

Our method combines variance-calibrated weight selection (CVR) with energy-compensated weight adjustment and is denoted as Ours (CVR+EC).
In addition, for all baselines, the proposed energy-compensated weight adjustment is applied as a post-pruning correction on top of the corresponding pruning masks, with the resulting methods llabeled as ``+EC'' (e.g., ``Wanda+EC'')

\paragraph{Calibration Data.}
For methods that require calibration data for mask construction (e.g., Wanda and SparseGPT), we follow the same protocol as Wanda~\citep{wanda}.
Specifically, we sample 128 sequences from the C4 dataset~\cite{raffel2023c4} to collect activation statistics.

The proposed method does not rely on additional calibration data beyond that used for mask construction.

\paragraph{Evaluation Metrics.}
We evaluate pruned models using perplexity on WikiText2~\cite{merity2016wikitext2} for language modeling performance and average zero-shot accuracy on a suite of downstream tasks, including BoolQ~\cite{clark2019boolq}, RTE~\cite{wang2019rte}, HellaSwag~\cite{zellers2019hellaswag}, WinoGrande~\cite{sakaguchi2019winogrande}, ARC~\cite{clark2018ARC}, and OBQA~\cite{mihaylov2018obqa}.
All evaluations are conducted without retraining or fine-tuning unless explicitly stated.

\begin{table*}[t]
\caption{Zero-shot accuracy (\%) under different sparsity settings. Higher is better.}
\label{tab:wikitext_zs_with_ec}
\centering
\begin{small}
\setlength{\tabcolsep}{4pt}
\renewcommand{\arraystretch}{1.15}
\begin{tabular}{l c c ccc cc cccc}
\toprule
\multirow{2}{*}{Method} & \multirow{2}{*}{Weight Update} & \multirow{2}{*}{Sparsity}
& \multicolumn{3}{c}{LLaMA-2}
& \multicolumn{2}{c}{LLaMA-3}
& \multicolumn{4}{c}{Qwen2.5} \\
\cmidrule(lr){4-6}\cmidrule(lr){7-8}\cmidrule(lr){9-12}
& & & 7B & 13B & 70B & 8B & 70B & 7B & 14B & 32B & 72B \\
\midrule
Dense & -- & 0
& 59.71 & 63.03 & 67.08 & 72.72 & 80.05 & 72.18 & 75.81 & 75.12 & 78.69 \\
\midrule

\rowcolor{gray!10}
Magnitude & No & 50\%
& 51.14 & 52.85 & 60.93 & 41.80 & 64.05 & 43.02 & 55.07 & 62.96 & 41.27 \\
\rowcolor{gray!10}
Magnitude+EC & Yes & 50\%
& \textbf{57.45} & \textbf{60.74} & \textbf{67.95} & \textbf{42.56} & \textbf{64.57} & \textbf{48.78} & \textbf{58.59} & \textbf{63.13} & \textbf{54.64} \\
\rowcolor{blue!10}
Wanda & No & 50\%
& 56.24 & 60.83 & 67.03 & 60.71 & \textbf{70.28} & 58.77 & 63.05 & 70.99 & \textbf{73.58} \\
\rowcolor{blue!10}
Wanda+EC & Yes & 50\%
& \textbf{61.35} & \textbf{64.51} & \textbf{71.02} & \textbf{62.60} & 65.80 & \textbf{67.14} & \textbf{68.78} & \textbf{71.26} & 73.45 \\
\rowcolor{gray!20}
Ours (CVR+EC) & Yes & 50\%
& 60.95 & 64.31 & 70.78 & 62.17 & 68.45 & \textbf{67.25} & \textbf{69.06} & 69.90 & \textbf{73.78} \\
\midrule

\rowcolor{gray!10}
Magnitude & No & 4:8
& 50.64 & 52.81 & 60.28 & 44.40 & 58.13 & 40.68 & 52.96 & 60.72 & 40.57 \\
\rowcolor{gray!10}
Magnitude+EC & Yes & 4:8
& \textbf{55.33} & \textbf{58.49} & \textbf{65.85} & \textbf{44.79} & \textbf{60.91} & \textbf{46.83} & \textbf{58.45} & \textbf{61.55} & \textbf{51.33} \\
\rowcolor{blue!10}
Wanda & No & 4:8
& 52.49 & 58.75 & 66.06 & 55.63 & \textbf{67.91} & 53.00 & 55.56 & \textbf{70.60} & \textbf{72.92} \\
\rowcolor{blue!10}
Wanda+EC & Yes & 4:8
& \textbf{59.27} & \textbf{62.48} & \textbf{70.30} & \textbf{55.94} & 58.60 & \textbf{66.17} & \textbf{67.05} & 70.39 & 72.90 \\
\rowcolor{gray!20}
Ours (CVR+EC) & Yes & 4:8
& \textbf{59.30} & 62.27 & \textbf{70.50} & \textbf{56.01} & 62.35 & 66.03 & \textbf{68.79} & 69.25 & \textbf{72.98} \\
\midrule

\rowcolor{gray!10}
Magnitude & No & 2:4
& 45.58 & 49.89 & 59.95 & 38.95 & \textbf{55.73} & 40.73 & 48.60 & 59.79 & 41.41 \\
\rowcolor{gray!10}
Magnitude+EC & Yes & 2:4
& \textbf{52.10} & \textbf{55.92} & \textbf{64.97} & \textbf{40.81} & 51.84 & \textbf{45.52} & \textbf{53.14} & \textbf{60.13} & \textbf{51.58} \\
\rowcolor{blue!10}
Wanda & No & 2:4
& 48.75 & 55.03 & 64.14 & 51.03 & \textbf{61.40} & 43.11 & 50.46 & \textbf{68.62} & 71.75 \\
\rowcolor{blue!10}
Wanda+EC & Yes & 2:4
& \textbf{56.77} & \textbf{61.38} & \textbf{68.37} & \textbf{52.78} & 49.23 & \textbf{60.64} & \textbf{63.08} & 67.89 & \textbf{71.77} \\
\rowcolor{gray!20}
Ours (CVR+EC) & Yes & 2:4
& 56.46 & \textbf{61.66} & \textbf{68.72} & 51.47 & 59.53 & \textbf{62.55} & \textbf{63.70} & 67.05 & 71.74 \\
\bottomrule
\end{tabular}
\end{small}
\end{table*}

\subsection{Language Modeling}
\label{sec:language_modeling}

We evaluate language modeling performance using perplexity on the WikiText2~\cite{merity2016wikitext2} dataset.
Table~1 reports results under both unstructured sparsity (50\%) and structured sparsity patterns (4:8 and 2:4) across different model families and scales.

Under 50\% unstructured sparsity, applying energy compensation consistently improves perplexity over the corresponding pruning baselines.
The proposed method achieves the lowest perplexity on all LLaMA-2 models and attains competitive performance on LLaMA-3 and Qwen2.5 models, often matching or improving upon magnitude-based and activation-based baselines.
These results indicate that post-pruning correction effectively mitigates the degradation introduced by unstructured pruning.

Similar trends are observed under structured sparsity.
For both 4:8 and 2:4 patterns, energy compensation reduces perplexity for magnitude-based pruning and Wanda across most models.
The proposed method maintains stable performance across architectures and sparsity patterns, achieving the best or second-best results in many configurations, particularly on LLaMA-2 and Qwen2.5 models.

Overall, these results suggest that part of the performance degradation induced by pruning may be attributed to distributional changes in the remaining weights, in addition to the choice of pruning masks.
Applying a lightweight statistical correction after pruning helps mitigate this effect and improves language modeling performance in many evaluated settings, without requiring retraining or reconstruction-based optimization.

\subsection{Zero-shot Evaluation}
\label{sec:zeroshot}

We evaluate pruned models on a suite of zero-shot downstream tasks following prior work~\citep{wanda}.
Table~\ref{tab:wikitext_zs_with_ec} reports the average accuracy across six tasks under different sparsity settings.

Zero-shot accuracy decreases with increasing sparsity across all methods and model families.
Applying energy compensation consistently improves performance over the corresponding pruning baselines, indicating that correcting pruning-induced scale distortion is important for preserving task-relevant representations.

Under 50\% unstructured sparsity, energy compensation gives clear gains for both magnitude-based pruning and Wanda across most models.
The proposed method achieves competitive accuracy across model sizes, matching or exceeding magnitude-based and activation-based baselines in many settings, particularly on LLaMA-2 and Qwen2.5 models.
Similar trends are observed under structured sparsity (4:8 and 2:4), where energy compensation improves zero-shot accuracy over the corresponding baselines.

In several settings, Wanda+EC performs comparably to CVR+EC, suggesting that zero-shot performance is primarily driven by post-pruning distributional correction rather than fine-grained differences in mask selection.

Taken together with the perplexity results, these findings indicate that post-pruning energy correction can effectively preserve task-relevant representations.
By reducing pruning-induced distributional distortion without additional training or reconstruction-based optimization, the proposed method provides a lightweight approach for maintaining zero-shot performance across different sparsity settings.

\begin{table}[t]
\caption{Runtime comparison for post-training pruning methods. Speedup is measured relative to SparseGPT.}
\label{tab:efficiency}
\centering
\begin{small}
\setlength{\tabcolsep}{6pt}
\renewcommand{\arraystretch}{1.15}
\begin{tabular}{lccc}
\toprule
Method & Weight Update  & Runtime (s)$\downarrow$ & Speedup $\uparrow$ \\
\midrule
Magnitude & \xmark  &0.11 & 274.7$\times$ \\
Wanda & \xmark  &4.92 & 6.31$\times$  \\
SparseGPT & \cmark  &31.07  & 1.00$\times$\\
\rowcolor{gray!10}
Ours & \cmark  &5.35 & 5.80$\times$ \\
\bottomrule
\end{tabular}
\end{small}
\end{table}




\subsection{Efficiency and Runtime Analysis}
\label{sec:efficiency}

We compare the runtime overhead of different post-training pruning methods.
Table~\ref{tab:efficiency} reports the runtime measured on a single linear layer of LLaMA-2-13B, focusing on the cost of pruning and post-pruning weight adjustment.

Magnitude pruning and Wanda have low runtime, as they only involve simple thresholding or importance-based masking without weight updates.
SparseGPT is included as a reference for reconstruction-based methods, which has significantly higher overhead due to second-order statistics computation and reconstruction-based updates.

The proposed method also adjusts weights after pruning, yet achieves runtime comparable to Wanda and requires no additional calibration beyond mask construction, indicating that the correction can be applied efficiently.

\subsection{Robustness to Unstructured Sparsity Levels}

We analyze robustness to high unstructured sparsity using perplexity on LLaMA-2-13B, with results shown in Table~4.
As sparsity increases, all methods experience performance degradation, though the degree of degradation varies across approaches.

We include SparseGPT as a representative reconstruction-based method that leverages second-order information to improve robustness under aggressive pruning.
Consistent with prior work, magnitude-based pruning and activation-based heuristics degrade rapidly beyond 60\% sparsity, exhibiting sharp increases in perplexity.
SparseGPT mitigates this degradation by explicitly updating weights during pruning, achieving improved stability compared to heuristic methods.

Across the evaluated sparsity range, the proposed method gives the most stable performance.
It achieves lower perplexity than both heuristic baselines and SparseGPT at high sparsity levels, despite relying only on lightweight statistical correction and avoiding second-order reconstruction.
These results indicate that correcting pruning-induced distributional distortion can improve robustness under high sparsity without the computational cost of reconstruction-based optimization.

\begin{table}[t]
\caption{Perplexity of different pruning methods across unstructured sparsity levels on LLaMA-2-13B. Lower is better.}
\label{tab:ablation_sparsity_llama2_13b}
\centering
\begin{small}
\setlength{\tabcolsep}{6pt}
\renewcommand{\arraystretch}{1.15}
\begin{tabular}{lccccc}
\toprule
Method & 10\% & 30\% & 50\% &60\% & 70\% \\
\midrule
Magnitude & 4.59 & 4.82 & 6.37 &11.22 & 275.23 \\
Wanda     & 4.60 & 4.79 & 5.59 &7.02 & 45.77 \\
SparseGPT & 4.59 & 4.78 & 5.63 &7.74 & 18.04  \\
\rowcolor{gray!10}
Ours & \textbf{4.57} & \textbf{4.72} & \textbf{5.32} &\textbf{6.33} & \textbf{10.52}\\
\bottomrule
\end{tabular}
\end{small}\vspace{-1em}
\end{table}


\subsection{Effect of Calibration Set Size}
\label{sec:calibration_size}

\begin{figure}[t]
  \centering
  \includegraphics[width=\columnwidth]{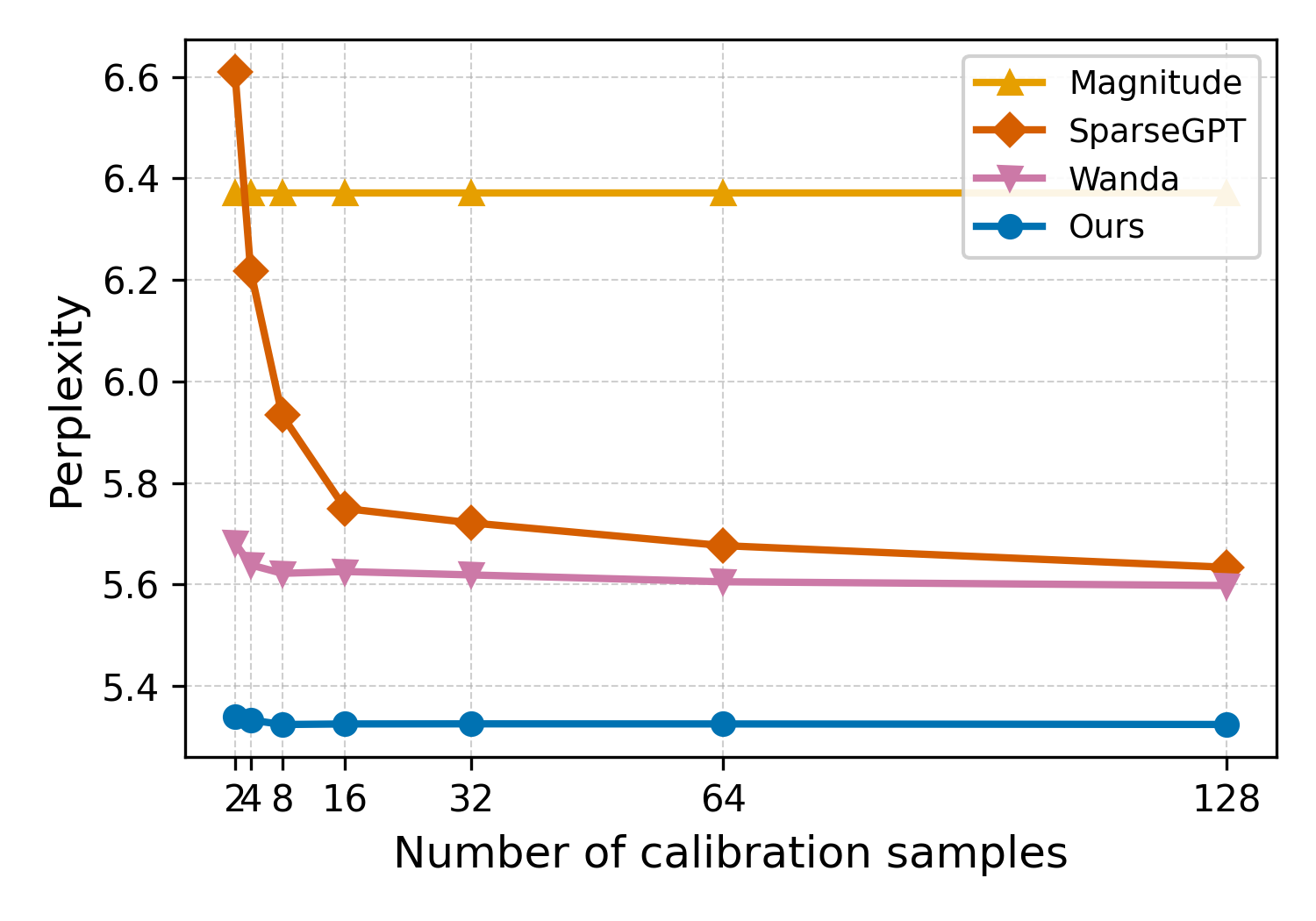}\vspace{-1em}
  \caption{
  Perplexity across calibration set sizes on LLaMA-2-13B with 50\% unstructured sparsity. Lower is better.}
  \label{fig:calib_size}\vspace{-1em}
\end{figure}

Figure~\ref{fig:calib_size} examines the effect of calibration set size under 50\% unstructured sparsity on LLaMA-2-13B.
Across all calibration sizes, the proposed method achieves lower perplexity than the compared baselines and remains stable even with very limited calibration data.

This indicates that the post-pruning correction relies primarily on intrinsic statistics of the pruned weights rather than fitting to the calibration set.
As a result, the method remains effective with a small amount of calibration data.

\subsection{Ablation Study}
\label{sec:ablation}

We conduct an ablation study to analyze the contribution of different components in the proposed statistical correction framework.
Results on LLaMA-2-7B and LLaMA-2-13B are reported in Table \ref{tab:ablation_clean}.

We start from Wanda as the baseline pruning method.
Adding variance-calibrated reweighting (CVR) consistently reduces perplexity, indicating that correcting activation-dominated importance estimates helps suppress unstable channels during pruning.

We then examine the effect of post-pruning energy compensation.
Applying column-wise energy compensation (EC$_{\mathrm{col}}$) further improves perplexity over CVR alone, suggesting that pruning introduces systematic energy loss along input dimensions.
Finally, adding row-wise energy compensation on top of column-wise correction (EC$_{\mathrm{row}}$) yields additional gains.

Overall, the results indicate that variance-calibrated selection and energy compensation address complementary sources of pruning-induced degradation, and that combining both components leads to the most effective post-training correction in this setting.

\section{Related Work}

\subsection{Post-Training Pruning for LLMs}

Post-training pruning is commonly used to reduce the size and inference cost of LLMs without additional training. Early approaches rely on metrics derived from weight magnitude~\cite{magnitude,han2016deepcompressioncompressingdeep,zhu2017prunepruneexploringefficacy,blalock2020stateneuralnetworkpruning,molchanov2019importanceestimationneuralnetwork}.
More recent methods such as Wanda~\cite{wanda} incorporate activation information \cite{molchanov2017pruningconvolutionalneuralnetworks,lee2019snipsingleshotnetworkpruning} to better capture input-dependent behavior,  where weight–activation statistics are used to guide pruning and improves over magnitude-based pruning metrics.

Another line of work formulates pruning as a reconstruction problem. Methods such as SparseGPT~\cite{sparsegpt} and related extensions~\cite{ma2023llmpruner,vanderouderaa2024llmsurgeon,frantar2023gptqaccurateposttrainingquantization,frantar2023optimalbraincompressionframework,singh2020woodfisherefficientsecondorderapproximation,hubara2021acceleratedsparseneuraltraining} update the remaining weights after pruning to reduce reconstruction error.
These methods typically rely on second-order information and have higher computational and memory cost.

Some other works take alternative approaches to modify the pruning process itself. DenoiseRotator~\cite{denoiserotator} applies orthogonal transformations before pruning to redistribute parameter importance.
RIA~\cite{ria} proposes a relative importance metric based on channel-wise normalization to mitigate scale imbalance during pruning. These methods mainly focus on how importance is defined or adjusted prior to pruning \cite{gale2019statesparsitydeepneural}.

\begin{table}[t]
\caption{Ablation study of statistical correction components on LLaMA-2-7B and LLaMA-2-13B.}
\label{tab:ablation_clean}
\centering
\small
\setlength{\tabcolsep}{8pt}
\renewcommand{\arraystretch}{1.15}
\begin{tabular}{l l cc}
\toprule
\textbf{Selection}  & \textbf{7B} & \textbf{13B} \\
\midrule
Wanda & 6.42 & 5.46 \\
+ CVR   & 6.32 & 5.43 \\
+ EC$_{\mathrm{col}}$ & 6.21 & 5.33 \\

+ EC$_{\mathrm{row}}$ & 6.19 & 5.32 \\
\bottomrule
\end{tabular}\vspace{-1em}
\end{table}

\subsection{Statistical Properties of Weights and Activations}

The statistical properties of weights and activations in LLMs have been studied in the context of model robustness and model compression \cite{bondarenko2021understandingovercomingchallengesefficient,Puccetti_2022}.
In quantization, prior work shows that activation outliers are common in transformer-based models \cite{kovaleva2021bertbustersoutlierdimensions, timkey2021barkbiteroguedimensions,wei2023outliersuppressionaccuratequantization} and can strongly affect compression performance~\cite{dettmers2022,xiao2023}.
These studies highlight the role of channel-wise heterogeneity and extreme activation values in determining sensitivity under compression.

\subsection{Post-Pruning Distribution Alignment}

Preserving the distribution of intermediate representations is important for maintaining model performance after model reduction \cite{ioffe2015batchnormalizationacceleratingdeep}.
Reconstruction-based pruning methods address this through explicit optimization objectives~\cite{sparsegpt,renda2020comparingrewindingfinetuningneural}.
In related areas such as quantization and domain adaptation, simpler approaches based on matching first- and second-order statistics of feature maps have been shown to be effective~\cite{li2016revisiting,sun2016deep}.
Such statistical alignment techniques rely on closed-form corrections rather than iterative optimization.

In the context of post-training pruning for LLMs, most prior work focuses on mask selection or reconstruction-based weight updates.
Lightweight post-pruning corrections based on feature distribution alignment have received comparatively less attention.

\section{Limitations}

The proposed approach relies on first-order statistical properties of weights and activations, which may be less effective under extreme distribution shift or highly non-stationary inputs.
In addition, while energy compensation improves stability under moderate to high sparsity, performance still degrades at extreme sparsity levels where model capacity is fundamentally limited.

Our evaluation focuses on decoder-only LLMs, which dominate current large-scale language modeling. Extending the analysis to encoder–decoder and multimodal architectures is an interesting direction for future work.

\section{Conclusion}
We presented a post-training pruning approach for LLMs based on lightweight statistical correction.
The method combines variance-aware importance calibration with post-pruning energy adjustment to address distributional distortions caused by weight sparsification.
Both components rely on first-order statistical quantities and do not require retraining or reconstruction-based optimization. 

Experimental results across multiple model families and sparsity settings show that the proposed approach improves pruning performance with minimal computational overhead.
These results suggest that simple statistical corrections can be effective for post-training pruning.
Future work includes extending this approach to joint pruning and quantization and evaluating its applicability to larger and more heterogeneous model architectures.

\section*{Impact Statement}

This paper presents work whose goal is to advance the field of machine learning by studying robustness and efficiency aspects of neural network pruning methods.
The proposed approach focuses on improving stability under high sparsity and is intended as a methodological contribution to model compression.

The techniques studied in this work operate at the level of model optimization and do not introduce new data sources, learning paradigms, or application domains.
As such, we do not anticipate societal or ethical impacts beyond those already associated with the broader development and use of large language models.

\nocite{langley00}

\bibliography{example_paper}

@misc{openai2024gpt4technicalreport,
      title={GPT-4 Technical Report}, 
      author={OpenAI},
      year={2024},
      eprint={2303.08774},
      archivePrefix={arXiv},
      primaryClass={cs.CL},
      url={https://arxiv.org/abs/2303.08774}, 
}

@misc{brown2020languagemodelsfewshotlearners,
      title={Language Models are Few-Shot Learners}, 
      author={OpenAI},
      year={2020},
      eprint={2005.14165},
      archivePrefix={arXiv},
      primaryClass={cs.CL},
      url={https://arxiv.org/abs/2005.14165}, 
}

@misc{llama,
      title={LLaMA: Open and Efficient Foundation Language Models}, 
      author={MetaAI},
      year={2023},
      eprint={2302.13971},
      archivePrefix={arXiv},
      primaryClass={cs.CL},
      url={https://arxiv.org/abs/2302.13971}, 
}

@misc{llama2,
      title={Llama 2: Open Foundation and Fine-Tuned Chat Models}, 
      author={MetaAI},
      year={2023},
      eprint={2307.09288},
      archivePrefix={arXiv},
      primaryClass={cs.CL},
      url={https://arxiv.org/abs/2307.09288}, 
}

@misc{llama3,
      title={The Llama 3 Herd of Models}, 
      author={MetaAI},
      year={2024},
      eprint={2407.21783},
      archivePrefix={arXiv},
      primaryClass={cs.AI},
      url={https://arxiv.org/abs/2407.21783}, 
}

@misc{wanda,
      title={A Simple and Effective Pruning Approach for Large Language Models}, 
      author={Mingjie Sun and Zhuang Liu and Anna Bair and J. Zico Kolter},
      year={2024},
      eprint={2306.11695},
      archivePrefix={arXiv},
      primaryClass={cs.CL},
      url={https://arxiv.org/abs/2306.11695}, 
}

@misc{sparsegpt,
      title={SparseGPT: Massive Language Models Can Be Accurately Pruned in One-Shot}, 
      author={Elias Frantar and Dan Alistarh},
      year={2023},
      eprint={2301.00774},
      archivePrefix={arXiv},
      primaryClass={cs.LG},
      url={https://arxiv.org/abs/2301.00774}, 
}

@misc{frantar2023gptqaccurateposttrainingquantization,
      title={GPTQ: Accurate Post-Training Quantization for Generative Pre-trained Transformers}, 
      author={Elias Frantar and Saleh Ashkboos and Torsten Hoefler and Dan Alistarh},
      year={2023},
      eprint={2210.17323},
      archivePrefix={arXiv},
      primaryClass={cs.LG},
      url={https://arxiv.org/abs/2210.17323}, 
}

@misc{frantar2023optimalbraincompressionframework,
      title={Optimal Brain Compression: A Framework for Accurate Post-Training Quantization and Pruning}, 
      author={Elias Frantar and Sidak Pal Singh and Dan Alistarh},
      year={2023},
      eprint={2208.11580},
      archivePrefix={arXiv},
      primaryClass={cs.LG},
      url={https://arxiv.org/abs/2208.11580}, 
}

@inproceedings{ria,
    title={Plug-and-Play: An Efficient Post-training Pruning Method for Large Language Models},
    author={Yingtao Zhang and Haoli Bai and Haokun Lin and Jialin Zhao and Lu Hou and Carlo Vittorio Cannistraci},
    booktitle={The Twelfth International Conference on Learning Representations},
    year={2024},
    url={https://openreview.net/forum?id=Tr0lPx9woF}
}

@misc{magnitude,
      title={Learning both Weights and Connections for Efficient Neural Networks}, 
      author={Song Han and Jeff Pool and John Tran and William J. Dally},
      year={2015},
      eprint={1506.02626},
      archivePrefix={arXiv},
      primaryClass={cs.NE},
      url={https://arxiv.org/abs/1506.02626}, 
}

@misc{denoiserotator,
      title={DenoiseRotator: Enhance Pruning Robustness for LLMs via Importance Concentration}, 
      author={Tianteng Gu and Bei Liu and Bo Xiao and Ke Zeng and Jiacheng Liu and Yanmin Qian},
      year={2025},
      eprint={2505.23049},
      archivePrefix={arXiv},
      primaryClass={cs.LG},
      url={https://arxiv.org/abs/2505.23049}, 
}

@misc{frankle2019lotterytickethypothesisfinding,
      title={The Lottery Ticket Hypothesis: Finding Sparse, Trainable Neural Networks}, 
      author={Jonathan Frankle and Michael Carbin},
      year={2019},
      eprint={1803.03635},
      archivePrefix={arXiv},
      primaryClass={cs.LG},
      url={https://arxiv.org/abs/1803.03635}, 
}

@misc{voita2023neuronslargelanguagemodels,
      title={Neurons in Large Language Models: Dead, N-gram, Positional}, 
      author={Elena Voita and Javier Ferrando and Christoforos Nalmpantis},
      year={2023},
      eprint={2309.04827},
      archivePrefix={arXiv},
      primaryClass={cs.CL},
      url={https://arxiv.org/abs/2309.04827}, 
}

@misc{sanh2020movementpruningadaptivesparsity,
      title={Movement Pruning: Adaptive Sparsity by Fine-Tuning}, 
      author={Victor Sanh and Thomas Wolf and Alexander M. Rush},
      year={2020},
      eprint={2005.07683},
      archivePrefix={arXiv},
      primaryClass={cs.CL},
      url={https://arxiv.org/abs/2005.07683}, 
}

@misc{mishra2021acceleratingsparsedeepneural,
      title={Accelerating Sparse Deep Neural Networks}, 
      author={Asit Mishra and Jorge Albericio Latorre and Jeff Pool and Darko Stosic and Dusan Stosic and Ganesh Venkatesh and Chong Yu and Paulius Micikevicius},
      year={2021},
      eprint={2104.08378},
      archivePrefix={arXiv},
      primaryClass={cs.LG},
      url={https://arxiv.org/abs/2104.08378}, 
}

@misc{xiao2023,
      title={SmoothQuant: Accurate and Efficient Post-Training Quantization for Large Language Models}, 
      author={Guangxuan Xiao and Ji Lin and Mickael Seznec and Hao Wu and Julien Demouth and Song Han},
      year={2024},
      eprint={2211.10438},
      archivePrefix={arXiv},
      primaryClass={cs.CL},
      url={https://arxiv.org/abs/2211.10438}, 
}

@inbook{LeCun1990OBD,
    author = {Cun, Yann Le and Denker, John S. and Solla, Sara A.},
    title = {Optimal brain damage},
    year = {1990},
    isbn = {1558601007},
    publisher = {Morgan Kaufmann Publishers Inc.},
    address = {San Francisco, CA, USA},
    booktitle = {Advances in Neural Information Processing Systems 2},
    pages = {598–605},
    numpages = {8}
}

@inproceedings{HassibiOBS1993,
      author={Hassibi, B. and Stork, D.G. and Wolff, G.J.},
      booktitle={IEEE International Conference on Neural Networks}, 
      title={Optimal Brain Surgeon and general network pruning}, 
      year={1993},
      volume={},
      number={},
      pages={293-299 vol.1},
      doi={10.1109/ICNN.1993.298572}
}

@misc{dettmers2022,
      title={LLM.int8(): 8-bit Matrix Multiplication for Transformers at Scale}, 
      author={Tim Dettmers and Mike Lewis and Younes Belkada and Luke Zettlemoyer},
      year={2022},
      eprint={2208.07339},
      archivePrefix={arXiv},
      primaryClass={cs.LG},
      url={https://arxiv.org/abs/2208.07339}, 
}

@misc{qwen25,
      title={Qwen2.5 Technical Report}, 
      author={Qwen},
      year={2025},
      eprint={2412.15115},
      archivePrefix={arXiv},
      primaryClass={cs.CL},
      url={https://arxiv.org/abs/2412.15115}, 
}

@misc{ma2023llmpruner,
      title={LLM-Pruner: On the Structural Pruning of Large Language Models}, 
      author={Xinyin Ma and Gongfan Fang and Xinchao Wang},
      year={2023},
      eprint={2305.11627},
      archivePrefix={arXiv},
      primaryClass={cs.CL},
      url={https://arxiv.org/abs/2305.11627}, 
}

@misc{vanderouderaa2024llmsurgeon,
      title={The LLM Surgeon}, 
      author={Tycho F. A. van der Ouderaa and Markus Nagel and Mart van Baalen and Yuki M. Asano and Tijmen Blankevoort},
      year={2024},
      eprint={2312.17244},
      archivePrefix={arXiv},
      primaryClass={cs.LG},
      url={https://arxiv.org/abs/2312.17244}, 
}

@misc{hubara2021acceleratedsparseneuraltraining,
      title={Accelerated Sparse Neural Training: A Provable and Efficient Method to Find N:M Transposable Masks}, 
      author={Itay Hubara and Brian Chmiel and Moshe Island and Ron Banner and Seffi Naor and Daniel Soudry},
      year={2021},
      eprint={2102.08124},
      archivePrefix={arXiv},
      primaryClass={cs.AI},
      url={https://arxiv.org/abs/2102.08124}, 
}

@misc{kovaleva2021bertbustersoutlierdimensions,
      title={BERT Busters: Outlier Dimensions that Disrupt Transformers}, 
      author={Olga Kovaleva and Saurabh Kulshreshtha and Anna Rogers and Anna Rumshisky},
      year={2021},
      eprint={2105.06990},
      archivePrefix={arXiv},
      primaryClass={cs.CL},
      url={https://arxiv.org/abs/2105.06990}, 
}

@misc{raffel2023c4,
      title={Exploring the Limits of Transfer Learning with a Unified Text-to-Text Transformer}, 
      author={Colin Raffel and Noam Shazeer and Adam Roberts and Katherine Lee and Sharan Narang and Michael Matena and Yanqi Zhou and Wei Li and Peter J. Liu},
      year={2023},
      eprint={1910.10683},
      archivePrefix={arXiv},
      primaryClass={cs.LG},
      url={https://arxiv.org/abs/1910.10683}, 
}

@misc{merity2016wikitext2,
      title={Pointer Sentinel Mixture Models}, 
      author={Stephen Merity and Caiming Xiong and James Bradbury and Richard Socher},
      year={2016},
      eprint={1609.07843},
      archivePrefix={arXiv},
      primaryClass={cs.CL},
      url={https://arxiv.org/abs/1609.07843}, 
}

@misc{clark2019boolq,
      title={BoolQ: Exploring the Surprising Difficulty of Natural Yes/No Questions}, 
      author={Christopher Clark and Kenton Lee and Ming-Wei Chang and Tom Kwiatkowski and Michael Collins and Kristina Toutanova},
      year={2019},
      eprint={1905.10044},
      archivePrefix={arXiv},
      primaryClass={cs.CL},
      url={https://arxiv.org/abs/1905.10044}, 
}

@misc{wang2019rte,
      title={GLUE: A Multi-Task Benchmark and Analysis Platform for Natural Language Understanding}, 
      author={Alex Wang and Amanpreet Singh and Julian Michael and Felix Hill and Omer Levy and Samuel R. Bowman},
      year={2019},
      eprint={1804.07461},
      archivePrefix={arXiv},
      primaryClass={cs.CL},
      url={https://arxiv.org/abs/1804.07461}, 
}

@misc{zellers2019hellaswag,
      title={HellaSwag: Can a Machine Really Finish Your Sentence?}, 
      author={Rowan Zellers and Ari Holtzman and Yonatan Bisk and Ali Farhadi and Yejin Choi},
      year={2019},
      eprint={1905.07830},
      archivePrefix={arXiv},
      primaryClass={cs.CL},
      url={https://arxiv.org/abs/1905.07830}, 
}

@misc{sakaguchi2019winogrande,
      title={WinoGrande: An Adversarial Winograd Schema Challenge at Scale}, 
      author={Keisuke Sakaguchi and Ronan Le Bras and Chandra Bhagavatula and Yejin Choi},
      year={2019},
      eprint={1907.10641},
      archivePrefix={arXiv},
      primaryClass={cs.CL},
      url={https://arxiv.org/abs/1907.10641}, 
}

@misc{clark2018ARC,
      title={Think you have Solved Question Answering? Try ARC, the AI2 Reasoning Challenge}, 
      author={Peter Clark and Isaac Cowhey and Oren Etzioni and Tushar Khot and Ashish Sabharwal and Carissa Schoenick and Oyvind Tafjord},
      year={2018},
      eprint={1803.05457},
      archivePrefix={arXiv},
      primaryClass={cs.AI},
      url={https://arxiv.org/abs/1803.05457}, 
}

@misc{mihaylov2018obqa,
      title={Can a Suit of Armor Conduct Electricity? A New Dataset for Open Book Question Answering}, 
      author={Todor Mihaylov and Peter Clark and Tushar Khot and Ashish Sabharwal},
      year={2018},
      eprint={1809.02789},
      archivePrefix={arXiv},
      primaryClass={cs.CL},
      url={https://arxiv.org/abs/1809.02789}, 
}

@misc{sun2016deep,
      title={Deep CORAL: Correlation Alignment for Deep Domain Adaptation}, 
      author={Baochen Sun and Kate Saenko},
      year={2016},
      eprint={1607.01719},
      archivePrefix={arXiv},
      primaryClass={cs.CV},
      url={https://arxiv.org/abs/1607.01719}, 
}

@misc{li2016revisiting,
      title={Revisiting Batch Normalization For Practical Domain Adaptation}, 
      author={Yanghao Li and Naiyan Wang and Jianping Shi and Jiaying Liu and Xiaodi Hou},
      year={2016},
      eprint={1603.04779},
      archivePrefix={arXiv},
      primaryClass={cs.CV},
      url={https://arxiv.org/abs/1603.04779}, 
}

@misc{bommarito2022gpttakesbarexam,
      title={GPT Takes the Bar Exam}, 
      author={Michael Bommarito II and Daniel Martin Katz},
      year={2022},
      eprint={2212.14402},
      archivePrefix={arXiv},
      primaryClass={cs.CL},
      url={https://arxiv.org/abs/2212.14402}, 
}

@misc{wei2022emergentabilitieslargelanguage,
      title={Emergent Abilities of Large Language Models}, 
      author={Jason Wei and Yi Tay and Rishi Bommasani and Colin Raffel and Barret Zoph and Sebastian Borgeaud and Dani Yogatama and Maarten Bosma and Denny Zhou and Donald Metzler and Ed H. Chi and Tatsunori Hashimoto and Oriol Vinyals and Percy Liang and Jeff Dean and William Fedus},
      year={2022},
      eprint={2206.07682},
      archivePrefix={arXiv},
      primaryClass={cs.CL},
      url={https://arxiv.org/abs/2206.07682}, 
}

@misc{bubeck2023sparksartificialgeneralintelligence,
      title={Sparks of Artificial General Intelligence: Early experiments with GPT-4}, 
      author={Sébastien Bubeck and Varun Chandrasekaran and Ronen Eldan and Johannes Gehrke and Eric Horvitz and Ece Kamar and Peter Lee and Yin Tat Lee and Yuanzhi Li and Scott Lundberg and Harsha Nori and Hamid Palangi and Marco Tulio Ribeiro and Yi Zhang},
      year={2023},
      eprint={2303.12712},
      archivePrefix={arXiv},
      primaryClass={cs.CL},
      url={https://arxiv.org/abs/2303.12712}, 
}

@misc{lin2024awqactivationawareweightquantization,
      title={AWQ: Activation-aware Weight Quantization for LLM Compression and Acceleration}, 
      author={Ji Lin and Jiaming Tang and Haotian Tang and Shang Yang and Wei-Ming Chen and Wei-Chen Wang and Guangxuan Xiao and Xingyu Dang and Chuang Gan and Song Han},
      year={2024},
      eprint={2306.00978},
      archivePrefix={arXiv},
      primaryClass={cs.CL},
      url={https://arxiv.org/abs/2306.00978}, 
}

@misc{hoffmann2022trainingcomputeoptimallargelanguage,
      title={Training Compute-Optimal Large Language Models}, 
      author={DeepMind},
      year={2022},
      eprint={2203.15556},
      archivePrefix={arXiv},
      primaryClass={cs.CL},
      url={https://arxiv.org/abs/2203.15556}, 
}

@misc{strubell2019energypolicyconsiderationsdeep,
      title={Energy and Policy Considerations for Deep Learning in NLP}, 
      author={Emma Strubell and Ananya Ganesh and Andrew McCallum},
      year={2019},
      eprint={1906.02243},
      archivePrefix={arXiv},
      primaryClass={cs.CL},
      url={https://arxiv.org/abs/1906.02243}, 
}

@misc{timkey2021barkbiteroguedimensions,
      title={All Bark and No Bite: Rogue Dimensions in Transformer Language Models Obscure Representational Quality}, 
      author={William Timkey and Marten van Schijndel},
      year={2021},
      eprint={2109.04404},
      archivePrefix={arXiv},
      primaryClass={cs.CL},
      url={https://arxiv.org/abs/2109.04404}, 
}

@misc{kaplan2020scalinglawsneurallanguage,
      title={Scaling Laws for Neural Language Models}, 
      author={Jared Kaplan and Sam McCandlish and Tom Henighan and Tom B. Brown and Benjamin Chess and Rewon Child and Scott Gray and Alec Radford and Jeffrey Wu and Dario Amodei},
      year={2020},
      eprint={2001.08361},
      archivePrefix={arXiv},
      primaryClass={cs.LG},
      url={https://arxiv.org/abs/2001.08361}, 
}

@misc{frankle2021pruningneuralnetworksinitialization,
      title={Pruning Neural Networks at Initialization: Why are We Missing the Mark?}, 
      author={Jonathan Frankle and Gintare Karolina Dziugaite and Daniel M. Roy and Michael Carbin},
      year={2021},
      eprint={2009.08576},
      archivePrefix={arXiv},
      primaryClass={cs.LG},
      url={https://arxiv.org/abs/2009.08576}, 
}

@misc{devlin2019bertpretrainingdeepbidirectional,
      title={BERT: Pre-training of Deep Bidirectional Transformers for Language Understanding}, 
      author={Jacob Devlin and Ming-Wei Chang and Kenton Lee and Kristina Toutanova},
      year={2019},
      eprint={1810.04805},
      archivePrefix={arXiv},
      primaryClass={cs.CL},
      url={https://arxiv.org/abs/1810.04805}, 
}

@misc{hendrycks2021measuringmassivemultitasklanguage,
      title={Measuring Massive Multitask Language Understanding}, 
      author={Dan Hendrycks and Collin Burns and Steven Basart and Andy Zou and Mantas Mazeika and Dawn Song and Jacob Steinhardt},
      year={2021},
      eprint={2009.03300},
      archivePrefix={arXiv},
      primaryClass={cs.CY},
      url={https://arxiv.org/abs/2009.03300}, 
}

@misc{gale2019statesparsitydeepneural,
      title={The State of Sparsity in Deep Neural Networks}, 
      author={Trevor Gale and Erich Elsen and Sara Hooker},
      year={2019},
      eprint={1902.09574},
      archivePrefix={arXiv},
      primaryClass={cs.LG},
      url={https://arxiv.org/abs/1902.09574}, 
}

@misc{zhu2017prunepruneexploringefficacy,
      title={To prune, or not to prune: exploring the efficacy of pruning for model compression}, 
      author={Michael Zhu and Suyog Gupta},
      year={2017},
      eprint={1710.01878},
      archivePrefix={arXiv},
      primaryClass={stat.ML},
      url={https://arxiv.org/abs/1710.01878}, 
}

@misc{han2016deepcompressioncompressingdeep,
      title={Deep Compression: Compressing Deep Neural Networks with Pruning, Trained Quantization and Huffman Coding}, 
      author={Song Han and Huizi Mao and William J. Dally},
      year={2016},
      eprint={1510.00149},
      archivePrefix={arXiv},
      primaryClass={cs.CV},
      url={https://arxiv.org/abs/1510.00149}, 
}

@misc{blalock2020stateneuralnetworkpruning,
      title={What is the State of Neural Network Pruning?}, 
      author={Davis Blalock and Jose Javier Gonzalez Ortiz and Jonathan Frankle and John Guttag},
      year={2020},
      eprint={2003.03033},
      archivePrefix={arXiv},
      primaryClass={cs.LG},
      url={https://arxiv.org/abs/2003.03033}, 
}

@misc{molchanov2017pruningconvolutionalneuralnetworks,
      title={Pruning Convolutional Neural Networks for Resource Efficient Inference}, 
      author={Pavlo Molchanov and Stephen Tyree and Tero Karras and Timo Aila and Jan Kautz},
      year={2017},
      eprint={1611.06440},
      archivePrefix={arXiv},
      primaryClass={cs.LG},
      url={https://arxiv.org/abs/1611.06440}, 
}

@misc{singh2020woodfisherefficientsecondorderapproximation,
      title={WoodFisher: Efficient Second-Order Approximation for Neural Network Compression}, 
      author={Sidak Pal Singh and Dan Alistarh},
      year={2020},
      eprint={2004.14340},
      archivePrefix={arXiv},
      primaryClass={cs.LG},
      url={https://arxiv.org/abs/2004.14340}, 
}

@misc{molchanov2019importanceestimationneuralnetwork,
      title={Importance Estimation for Neural Network Pruning}, 
      author={Pavlo Molchanov and Arun Mallya and Stephen Tyree and Iuri Frosio and Jan Kautz},
      year={2019},
      eprint={1906.10771},
      archivePrefix={arXiv},
      primaryClass={cs.LG},
      url={https://arxiv.org/abs/1906.10771}, 
}

@misc{lee2019snipsingleshotnetworkpruning,
      title={SNIP: Single-shot Network Pruning based on Connection Sensitivity}, 
      author={Namhoon Lee and Thalaiyasingam Ajanthan and Philip H. S. Torr},
      year={2019},
      eprint={1810.02340},
      archivePrefix={arXiv},
      primaryClass={cs.CV},
      url={https://arxiv.org/abs/1810.02340}, 
}

@inproceedings{Puccetti_2022,
   title={Outlier Dimensions that Disrupt Transformers are Driven by Frequency},
   url={http://dx.doi.org/10.18653/v1/2022.findings-emnlp.93},
   DOI={10.18653/v1/2022.findings-emnlp.93},
   booktitle={Findings of the Association for Computational Linguistics: EMNLP 2022},
   publisher={Association for Computational Linguistics},
   author={Puccetti, Giovanni and Rogers, Anna and Drozd, Aleksandr and Dell’Orletta, Felice},
   year={2022},
   pages={1286–1304} 
}

@misc{bondarenko2021understandingovercomingchallengesefficient,
      title={Understanding and Overcoming the Challenges of Efficient Transformer Quantization}, 
      author={Yelysei Bondarenko and Markus Nagel and Tijmen Blankevoort},
      year={2021},
      eprint={2109.12948},
      archivePrefix={arXiv},
      primaryClass={cs.LG},
      url={https://arxiv.org/abs/2109.12948}, 
}

@misc{wei2023outliersuppressionaccuratequantization,
      title={Outlier Suppression+: Accurate quantization of large language models by equivalent and optimal shifting and scaling}, 
      author={Xiuying Wei and Yunchen Zhang and Yuhang Li and Xiangguo Zhang and Ruihao Gong and Jinyang Guo and Xianglong Liu},
      year={2023},
      eprint={2304.09145},
      archivePrefix={arXiv},
      primaryClass={cs.CL},
      url={https://arxiv.org/abs/2304.09145}, 
}

@misc{ioffe2015batchnormalizationacceleratingdeep,
      title={Batch Normalization: Accelerating Deep Network Training by Reducing Internal Covariate Shift}, 
      author={Sergey Ioffe and Christian Szegedy},
      year={2015},
      eprint={1502.03167},
      archivePrefix={arXiv},
      primaryClass={cs.LG},
      url={https://arxiv.org/abs/1502.03167}, 
}

@misc{renda2020comparingrewindingfinetuningneural,
      title={Comparing Rewinding and Fine-tuning in Neural Network Pruning}, 
      author={Alex Renda and Jonathan Frankle and Michael Carbin},
      year={2020},
      eprint={2003.02389},
      archivePrefix={arXiv},
      primaryClass={cs.LG},
      url={https://arxiv.org/abs/2003.02389}, 
}
\bibliographystyle{icml2026}

\newpage



\end{document}